\documentclass[letterpaper, 10 pt, conference]{ieeeconf}  

\IEEEoverridecommandlockouts                              

\usepackage{graphicx} 
\usepackage{amsmath} 
\usepackage{amssymb}  
\usepackage{wrapfig}
\usepackage{algorithm}
\usepackage{algpseudocode}
\usepackage{subcaption}
\usepackage{booktabs}
\usepackage{balance}
\usepackage{tikz}

\newcommand\copyrighttext{%
  \footnotesize \textcopyright 2025 IEEE. Personal use of this material is permitted.
  Permission from IEEE must be obtained for all other uses, in any current or future
  media, including reprinting/republishing this material for advertising or promotional
  purposes, creating new collective works, for resale or redistribution to servers or
  lists, or reuse of any copyrighted component of this work in other works.
  }
\newcommand\copyrightnotice{%
\begin{tikzpicture}[remember picture,overlay]
\node[anchor=south,yshift=10pt] at (current page.south) {\fbox{\parbox{\dimexpr\textwidth-\fboxsep-\fboxrule\relax}{\copyrighttext}}};
\end{tikzpicture}%
}

\newcommand\acceptancetext{%
\centering
  \footnotesize Accepted for publication at the IEEE International Conference on Robotics and Automation (ICRA), 2025
  }
\newcommand\acceptancenotice{%
\begin{tikzpicture}[remember picture,overlay]
\node[anchor=north,yshift=-10pt] at (current page.north)
{\parbox{\dimexpr\textwidth-\fboxsep-\fboxrule\relax}
{\acceptancetext}};
\end{tikzpicture}%
} 

\title{\LARGE \bf
Incremental Few-Shot Adaptation for Non-Prehensile Object Manipulation using Parallelizable Physics Simulators
}

\author{Fabian Baumeister$^{1}$, Lukas Mack$^{1, 2}$, and Joerg Stueckler$^{1,2}$
\thanks{*This work has been supported by Max Planck Society, Cyber Valley, the Cyber Valley Research Fund project CyVy-RF-2019-06, Hightech Agenda Bayern, and Deutsche Forschungsgemeinschaft (DFG) project no. 466606396 (STU 771/1-1).
We thank Felix Grueninger (MPI-IS) for building the robot end-effector used in our experiments.
  }
\thanks{$^{1}$All authors are with Embodied Vision Group, Max Planck Institute for Intelligent Systems, 72076 Tübingen, Germany
}%
\thanks{$^{2}$Lukas Mack and Joerg Stueckler are with Intelligent Perception in Technical Systems Group, University of Augsburg, 86159 Augsburg, Germany
{\tt\small firstname.lastname@uni-a.de}
}%
}

\begin{document}

\maketitle
\thispagestyle{empty}
\pagestyle{empty}
\acceptancenotice
\copyrightnotice 
\vspace{-\baselineskip} 

\begin{abstract}
Few-shot adaptation is an important capability for intelligent robots that perform tasks in open-world settings such as everyday environments or flexible production.
In this paper, we propose a novel approach for non-prehensile manipulation which incrementally adapts a physics-based dynamics model for model-predictive control (MPC).
The model prediction is aligned with a few examples of robot-object interactions collected with the MPC.
This is achieved by using a parallelizable rigid-body physics simulation as dynamic world model and sampling-based optimization of the model parameters.
In turn, the optimized dynamics model can be used for MPC using efficient sampling-based optimization.
We evaluate our few-shot adaptation approach in object pushing experiments in simulation and with a real robot.  
\end{abstract}

\section{INTRODUCTION}
Robots that perform object manipulation tasks in open-world scenarios such as everyday environments or in flexible production need the ability to quickly adapt to novel tasks and objects.
Model-based learning approaches that learn an action-conditional dynamics model of the environment are a promising direction to combine learning with optimization-based control and planning.
In recent years, several of such approaches have been proposed (e.g.,~\cite{finn2017_deepvisualforesight,sancaktar2022_ceeus,nagabandi2019_meta_mbrl}).
Analytical physics simulation is a capable approach for simulating multi-object dynamics.
Some methods have been recently evaluating the usage of such simulators for model-predictive control~\cite{howell2022_predictivesampling,pezzato2023_mppiisaac}, fueled by the development of efficient parallelizable simulators.
However, the shortcoming of these simulators is that realistic settings for the inherent parameters are difficult to achieve.

In this paper, we propose a novel real2sim2real approach which adapts a physics simulator to mimick real-world robot-object interaction behavior by optimizing the parameters of the simulation to better align with interaction data collected incrementally in the real world.
The optimized parameters are in turn used to plan goal-directed robot interactions with the object in real-time.
We build on the parallelizable rigid-body physics simulation of MuJoCo~\cite{todorov2012_mujoco}.
Parameter optimization and planning for model-predictive control are efficiently achieved using the cross-entropy method (CEM~\cite{rubinstein2004_cem}) for sampling-based optimization.
To avoid planning high frequency (e.g., 1\,kHz) robot controls, we propose to plan keypoints of minimum snap trajectories at much lower rates (e.g., 10\,Hz).
Our learning approach is few-shot because it can learn incrementally from a few rollouts of robot-object interactions.
In experiments, we analyze our few-shot learning approach in simulation and with a real robot for non-prehensile object manipulation, i.e., object pushing.
We demonstrate that our approach can optimize dynamics model parameters to better align with actual trajectory replays in simulation and can improve task execution performance with the real robot for offset parameter initializations.

In summary, our contributions are:
(1) We propose an incremental few-shot adaptation approach for non-prehensile object manipulation using CEM with physics simulation as dynamics model. The simulation parameters are adapted to better align the simulation with actual robot-object interaction rollouts in a replay buffer collected during task execution. 
(2) We plan keypoints of smooth minimum snap trajectories to facilitate real-time control using the optimized dynamics model.
(3) We demonstrate and assess our approach for non-prehensile object manipulation, i.e., object pushing, in simulation and with a real robot.

\section{RELATED WORK}

\begin{figure*}[tp!]
\centering
\vspace*{1.2ex}
\includegraphics[width=0.95\textwidth]{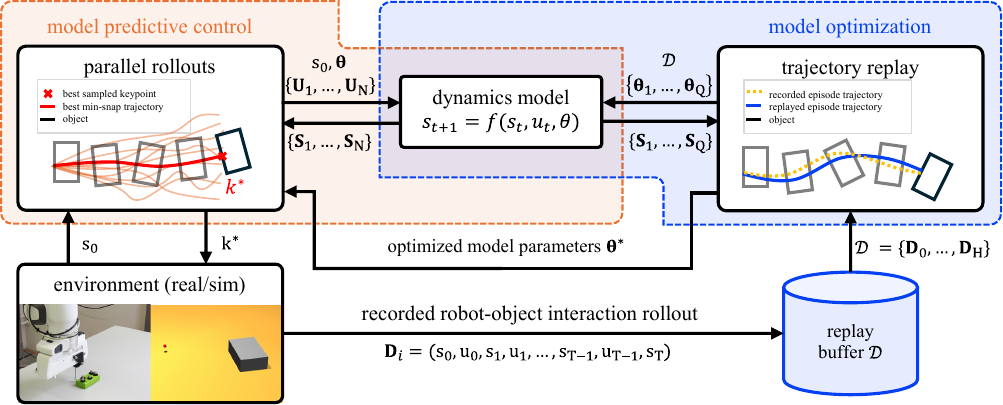}
\caption{Our approach is twofold: Rigid-body physics simulation is utilized as a dynamics model in model predictive control (MPC) to plan target keypoints for a minimum snap Cartesian trajectory.
To better match the dynamics model to the environment (e.g. real world) and thereby improve the performance of the MPC, observed robot-object interaction rollouts are consecutively stored in a replay buffer and used to optimize a subset of model parameters. Parameter optimization is iteratively executed by replaying the current buffer of trajectories and matching the resulting behavior to the ground truth. The optimized set of parameters $\boldsymbol{\theta}^*$ is then used by MPC to execute tasks. Environments: Real robot experiment setup (left) including end-effector sphere attachment and objects with motion capture markers; dynamics model/simulation setup (right) realizing robot end-effector (red sphere) and object (gray cuboid) as shape primitives.
}
\label{fig:method}
\end{figure*}

Learning dynamics models plays a key role for control with the learned models (e.g.,~\cite{finn2017_deepvisualforesight,sancaktar2022_ceeus,nagabandi2019_meta_mbrl}) or for model-based reinforcement learning (e.g.,~\cite{hafner2020_dreamer}).
Our approach can be attributed to the former class of approaches, but fits parameters of an analytical physics simulator instead of a learning-based model for fast adaptation for real-world object manipulation.
Several approaches employ model-predictive control for non-prehensile manipulation such as object pushing, using specific analytical models of object pushing effects (e.g.,~\cite{lynch1992_mechanicsofpushing,mason2001_mechanicsmanip,peshkin1988_workpiecemotion,moura2022_mpcc}). 
However, the approaches need accurate knowledge of object properties such as friction, mass, and inertia.  
Arruda et al.~\cite{arruda2017uncertainty} learn the dynamics of object pushing from motion capture data using an ensemble of Gaussian mixture models (GMMs~\cite{bishop2007_prml}) to avoid tedious parameter calibration and insufficiencies of analytical models.
The model is used in model-predictive path integral control (MPPI~\cite{kappen2005_mppi,theodorou2015_mppi}) to plan control actions to reach goals for the object.
The method is extended in~\cite{mathew2019_onlinelearning} to adapt the GMM online from interaction data.
Cong et al.~\cite{cong2020_selfadapting} propose to use a recurrent neural network as forward model trained on simulation data of various objects and its efficient integration into MPPI for control.
Video prediction models like in Finn et al.~\cite{finn2017_deepvisualforesight} learn forward dynamics models which encode the state of the environment in a neural latent space without explicit 3D object perception.
The model is used for object pushing using the cross entropy method~\cite{rubinstein2004_cem} for model-predictive control.
For real-time MPC with learned models, Bhardwaj et al.~\cite{bhardwaj2021_storm} propose an implementation which samples noisy control sequences for MPPI in a systematic way. 
To combine learning and analytical modeling, Hogan et al.~\cite{hogan2020_reactivepushing} propose to formulate object pushing by a mixed-integer MPC formulation using a quasi-static approximation of the robot-object interaction dynamics and propose to learn a hybrid model that combines analytical modeling and learning.
The mixed-integer optimization involves calculating the gradients of the dynamics model and can switch between different interaction models such as sticking and sliding.
Kloss et al.~\cite{kloss2020_contactreasoning} estimate the object pose on a plane from RGB-D images and filter physical state and friction parameters using an analytical model of pushing interactions. 
From the inferred parameters, the outcome of straight pushes on the object can be assessed using the analytical model for action planning.
They also learn an affordance model using CNNs and compare it with the analytical approach.  

In recent years, gradient-free sampling-based control methods have been proposed which use parallelizable physics simulators such as MuJoCo~\cite{todorov2012_mujoco} or Isaac Gym~\cite{makoviychuk2021_isaacgym} as forward models~\cite{howell2022_predictivesampling,pezzato2023_mppiisaac}. 
However, these approaches do not consider adapting the physics parameters to observations like our approach.
Some approaches learn physics-based models from robot-object interactions~\cite{sancaktar2022_ceeus} or optimize physics simulators with contacts and friction by making them differentiable~\cite{qiao2020_scalablediffphysics,strecke2021_diffsdfsim,kandukuri2023_ekfphys}. 
Related to our approach, \cite{barcelos2021_dust,AbrahamHRLMF20} extend MPPI~\cite{williams2017_mppi} to adapt the parameters of analytical dynamics models online.
DuST~\cite{barcelos2021_dust} collects state transition experience to infer parameters online and demonstrates the approach for mobile robot control.
Abraham et al.~\cite{AbrahamHRLMF20} adapt the single-step prediction of the dynamics model to the observed state transition in each control time step and demonstrate pushing and rotating a cube with a robot arm.
Chebotar et al.~\cite{ChebotarHMMIRF19} optimize simulation parameters during model-free RL using relative entropy policy search.
MACE~\cite{KrupnikSJT23} optimizes the parameters of a deep generative model for determining inverse kinematics of a robot arm that avoids obstacles.
Closely related is ASID~\cite{memmel2024_asid}, which trains an exploration policy to collect interactions from which the simulation parameters can be optimized using CEM~\cite{rubinstein1999_cem} by aligning simulated object trajectories with actual ones.
The method is demonstrated for planning the impact velocities for pushing the object at a goal distance for a real robot.
Differently to the above methods, our approach efficiently plans keypoints of smooth minimum snap trajectories and optimizes the dynamics model parameters alternatingly on a replay buffer of rollouts.
We demonstrate our approach for non-prehensile object manipulation in simulation and with a real robot.
The rollouts are collected incrementally using a goal-directed task objective.

\section{METHOD}

A high-level overview of the proposed approach is given in Fig. \ref{fig:method}. 
Our approach consists of a model-predictive control (MPC) component which performs tasks using the physics simulation as dynamics model of the environment and a task objective function.
The MPC plans keypoints of a minimum snap trajectory that is executed by the robot end-effector.
This way, the MPC is not required to plan low-level motion controls at 1\,kHz but instead plans smooth interactions at much lower rates (e.g., 10\,Hz).
While performing an object pushing task, robot-object-interaction data is collected in the environment and added to a replay buffer. 
Several parameters of the dynamics model are optimized such that the replayed task matches the actual real-world/simulation task execution better.
By doing so, subsequent runs will perform the task with improved parameters.

\subsection{Dynamic World Model}
To model the intricate dynamics of pushing an object through contact, we use the rigid-body physics simulator MuJoCo~\cite{todorov2012_mujoco} as dynamics model.
The dynamics model maps previous states~$\mathbf{s}_{t-1}$ and applied actions~$\mathbf{u}_t$ (e.g., end-effector motion) to a prediction of the next state~$\mathbf{s}_t = f( \mathbf{s}_{t-1}, \mathbf{u}_t, \boldsymbol{\theta} )$ according to physics-based rigid-body dynamics governed by~$P$ parameters~$\boldsymbol{\theta} \in \mathbb{R}^{P}$.
Inside the simulation, we model the robot end-effector and the~$O$ objects as shape primitives (sphere and cuboids, respectively).
The state $\mathbf{s}_{t}=(\mathbf{q}_{r}, \mathbf{q}_{o,1}, \ldots, \mathbf{q}_{o,O}, \mathbf{v}_{r}, \mathbf{v}_{o,1}, \ldots, \mathbf{v}_{o,O} )$ contains the poses and velocities of all rigid bodies, including the pose of the robot end-effector (modeled as a sphere) $\mathbf{q}_{r}(t) \in \mathbb{R}^3 \times \mathcal{S}^3$, the poses of each object~$\mathbf{q}_{o,k}(t) \in \mathbb{R}^3 \times \mathcal{S}^3$ and the velocities $\mathbf{v}_{r}(t), \mathbf{v}_{o,k}(t) \in \mathbb{R}^6$ where $k \in \{ 1, \ldots, O \}$.
The initial state $\mathbf{s}_0$ of the model is synchronized to the environment ahead of every MPC execution.
As the rigid-body physics simulation is only an approximation to the real world and corresponding friction parameters are difficult to obtain, tuning the physics parameters of the simulation
to best match the real world plays a key role for successful planning with the model.
Focusing on contact dynamics, the selected subset of simulation parameters of interest includes the object's sliding~$\theta_{s}$, torsional~$\theta_{t}$ and rolling~$\theta_{r}$ friction coefficients, as well as the end-effector sphere mass~$\theta_m$ (assuming the object mass is known). 
To feature parallel trajectory rollouts for optimization on CPU, we make use of MuJoCo's thread-safe C++ API.

\subsection{MPC for Non-Prehensile Manipulation}\label{sec:mpcnonprehense}
We use the dynamics model in model-predictive control~(MPC) to implement non-prehensile manipulation, i.e., object pushing.
MPC is achieved by efficient sampling-based optimization using the iCEM method~\cite{PinneriEtAl2020:iCEM}.
In order to alleviate the difficulty of generating high-frequency controls matching the robot's control rate (1 kHz) and to generate smooth trajectories, we parametrize the MPC actions as keypoints interpolated by minimum snap robot end-effector trajectories~\cite{richter2013_minsnap}. This allows for a much slower MPC rate (e.g., 10 Hz).
Prior to every control iteration the MPC model state is synchronized with the current environment state~$\mathbf{s}_{0}$.
Given the initial state~$\mathbf{s}_{0}$ a sequence of~$M$ keypoints $\mathbf{K} = \left( \mathbf{k}_1, \ldots, \mathbf{k}_M \right) \in \mathbb{R}^{M \times d}$ ($M=2$ in our experiments) is iteratively optimized using iCEM, where~$d$ is the dimensionality of the keypoints ($d=4$ in our experiments for 2D positions and velocities in the horizontal plane).
In each iteration~$l$, a population of~$N$ keypoint sequences $\mathbf{K}_i$ is drawn by sampling each keypoint $\mathbf{k}_{i,j}$ from~$\mathcal{N}(\boldsymbol{\mu}_{l,j}, \sigma_{l,j}^{2} \mathbf{I})$, where~$\mathbf{I}$ is the identity matrix.
Mean~$\boldsymbol{\mu}_{l,j}$ and standard deviation~$\sigma_{l,j}^{2}$ are set to predefined values in the first iteration.
In later iterations, they are determined from the elite set of keypoint sequences, i.e., the $p$ percent best evaluated sequences according to a cost function which is introduced later (we use $p=0.2$ in our experiments). 

An action sequence $\mathbf{U}_{i} = \left( \mathbf{u}_{i,1}, \ldots, \mathbf{u}_{i,MT} \right)$ is computed for every keypoint sequence, where~$T$ is the duration of the minimum snap trajectory for each keypoint. 
The resulting set of candidate state trajectories $\{ \mathbf{S}_{1},...,\mathbf{S}_{N} \} $ is simulated in parallel using MuJoCo's thread-safe C++ API. 
Each candidate is evaluated by the cost function 
$\mathcal{L}(\mathbf{S}_{i}) = \mathcal{L}_{\mathit{pos}}(\mathbf{s}_{i,MT}) 
    + \lambda_1 \mathcal{L}_{\mathit{rot}}(\mathbf{s}_{i,MT}) 
    + \lambda_2 \mathcal{L}_{\mathit{prox}}(\mathbf{s}_{i,MT}) 
    + \lambda_3 \mathcal{L}_{\mathit{acc}}(\mathbf{S}_{i})$,
for its
rollout state trajectory $\mathbf{S}_i$, with~$\mathbf{s}_{i,MT}$ denoting the final state, where~$\lambda_1, \lambda_2, \lambda_3$ are weighting parameters.
The positional cost term
    $\mathcal{L}_{\mathit{pos}}(\mathbf{s}_{i,MT}) = \left \| \mathbf{y}_{o,MT} - \mathbf{y}_o^{\ast} \right \|_2^{2}$
aims to minimize the distance between the final object 2D position~$\mathbf{y}_{o,MT}$ and a target position $\mathbf{y}_o^{\ast}$ in the horizontal plane.
Additionally, the rotational cost term
    $\mathcal{L}_{\mathit{rot}}(\mathbf{s}_{i,MT}) = (\sin{(\phi_{o,MT} - \hat{\phi}_{o,MT})})^2$
aims to align the final object yaw rotation~$\phi_{o,MT}$ to the target yaw rotation~$\hat{\phi}_{o,MT}$ invariant to the symmetry of the cuboids. 
In case the robot's end-effector is far away from the object, it does not have direct control over the object and sampling trajectories which interact with the object can be unlikely.
Hence, none of the candidate trajectories might contribute to the minimization of~$\mathcal{L}_{\mathit{goal}}(\mathbf{s}_{i,MT})$. 
Therefore, we introduce
$\mathcal{L}_{\mathit{prox}}(\mathbf{s}_{i,MT}) = \left \| \mathbf{y}_{o,MT} - \mathbf{y}_{r,MT} \right \|^{2}_2$
which favors proximity of the robot's end-effector position $\mathbf{y}_{r,MT}$ to the object at the end of the rollout.
We also include a control cost term
    $\mathcal{L}_{\mathit{acc}}(\mathbf{S}_{i}) =  \frac{1}{MT} \sum_{t=0}^{MT} \left \| \mathbf{a}_{o,t} \right \|^{2}_2$
to reduce the end-effector accelerations $\mathbf{a}_{o,i}$ within the trajectory candidate.

Given the current environment state~$\mathbf{s}_{0}$ and the first keypoint~$\mathbf{k}_{0}^{*}$ of the best performing candidate keypoint sequence~$\mathbf{K}^{*}$ according to $\mathcal{L}(\mathbf{S}_i)$,
we compute the minimum snap end-effector trajectory that is executed by the robot. 
A Cartesian position controller executes the new trajectory until the next keypoint is generated by the MPC.
Due to the minimum snap trajectory formulation, the position controller can switch to the new keypoint while the trajectory generation smoothly transits to the new keypoint.

We define the terminal conditions for the task as
    $\mathcal{L}_{\mathit{pos}}(\mathbf{s}_{i}(t))
    + \lambda_1 \mathcal{L}_{\mathit{rot}}(\mathbf{s}_{i}(t)) < \epsilon
    \wedge \, \quad t < \tau$,
where~$\mathbf{s}_i(t)$ is the actual state at the current execution time~$t$ (containing the current object and end-effector poses),~$\epsilon$ is the loss threshold, and~$\tau$ is the execution time limit.
The process is repeated until either of these conditions is met. 
If and only if the task terminates by satisfying the first condition, the task execution is considered to be \emph{successful}.

\subsection{Dynamics Model Parameter Optimization}

MPC performance depends on model quality.
We thus propose a method to adapt parameters of the dynamics model to decrease the mismatch between the model and real-world robot-object interactions which are observed during task execution. 
To this end, we replay the robot controls which were recorded during task execution. 
We make use of sampling-based optimization (CEM) to iteratively adapt model parameters. 
In simulation we achieve end-effector control by implementing a PID controller, applying external force onto the sphere. 
These controls can directly be replayed to optimize dynamics model parameters and decrease model mismatch.
On the real robot, we do not use noisy force measurements, but instead replay end-effector velocities.

In each optimization iteration~$l$, a population of~$Q$ parameters $\boldsymbol{\Theta} = \left\{ \boldsymbol{\theta}_{l,1}, \ldots, \boldsymbol{\theta}_{l,Q} \right\}$ is drawn from $\mathcal{N}(\boldsymbol{\nu}_l, \xi_l^{2} \mathbf{I})$, where~$\boldsymbol{\nu}_l \in \mathbb{R}^d$ is the mean and $\xi_l \in \mathbb{R}$ is the standard deviation of the normal distribution. 
In the first iteration, we sample from a prior parameter distribution with mean $\mathbf{\mu}_0 \in \mathbb{R}^{P}$ and standard deviation $\mathbf{\sigma}_0$. 
In subsequent iterations, the mean and standard deviation is determined by the distribution of the previous elite set (also using $p=0.2$ in our experiments). 
Let~$\mathcal{D}$ be a set of~$N$ rollouts $\mathbf{D}_i = \left( \mathbf{s}_{i,0}, \mathbf{u}_{i,0}, \ldots, \mathbf{s}_{i,T}, \mathbf{u}_{i,T} \right)$ in the environment with time duration~$T$.
For each candidate parameter setting, the robot controls of each rollout~$i$ are replayed. 
The resulting robot end-effector and object trajectories are assessed using the cost function
    $\mathcal{L}(\boldsymbol{\theta}) := \mathcal{L}_{\mathit{pos}, \mathit{o}} + \kappa_1 \mathcal{L}_{\mathit{pos}, \mathit{r}} + \kappa_2  \mathcal{L}_{\mathit{rot}, \mathit{o}}$, where
    $\mathcal{L}_{\mathit{pos}, \mathit{o}} := \sum_{i=1}^{N} \sum_{t=1}^{T_i} \left \| \mathbf{y}_{\mathit{o},i,t} - \widehat{\mathbf{y}}_{\mathit{o},i,t} \right \|_2^{2}$,
    $\mathcal{L}_{\mathit{pos}, \mathit{r}} := \sum_{i=1}^{N} \sum_{t=1}^{T_i} \left \| \mathbf{y}_{\mathit{r},i,t} - \widehat{\mathbf{y}}_{\mathit{r},i,t} \right \|_2^{2}$,
    $\mathcal{L}_{\mathit{rot}, \mathit{o}} :=  \sum_{i=1}^{N} \sum_{t=1}^{T_i} \arccos\left( \mathbf{v}\left(\phi_{\mathit{o},i,t}\right)^\top \, \mathbf{v}\left(\widehat{\phi}_{\mathit{o},i,t}\right) \right)^2$,
where~$\kappa_1,\kappa_2$ are weighting factors,~$T_i$ is the number of discrete steps of the respective trajectory,~$\mathbf{v}(\phi) := \left( \cos\left( \phi \right), \sin\left( \phi \right) \right)^\top$,~$\mathbf{y}_{\mathit{o},\ast}, \mathbf{y}_{\mathit{r},\ast}, \phi_{\mathit{o},\ast}$ are predicted positions and orientation angles, and~$\widehat{\mathbf{y}}_{\mathit{o},\ast}, \widehat{\mathbf{y}}_{\mathit{r},\ast}, \widehat{\phi}_{\mathit{o},\ast}$ are the actual states and angles for each rollout. 
A total of~$C$ CEM iterations is executed to optimize the parameters for a replay buffer. 
This optimization scheme enables few-shot learning by successively optimizing the dynamics model parameters after each MPC-based execution of an object manipulation tasks.
Once a task terminates in iteration $i$, the performed robot-object interaction rollout $\mathbf{D}_i$ is included into the replay buffer $\mathcal{D} \leftarrow \mathcal{D} \cup \left\{ \mathbf{D}_i \right\}$.
The parameters are optimized using the current replay buffer and the optimized parameters $\boldsymbol{\theta}_i^{*}$ are used for the next task execution.

\section{EXPERIMENTS}
We evaluate our method both in simulation and on a real robotic setup. 
The simulation allows for assessing ground-truth states and physical parameters and is used to evaluate the accuracy of our method in estimating the dynamics model parameters. 
Furthermore, we demonstrate real-time capability of our proposed sampling based MPC approach by carrying out a non-prehensile object manipulation task on a 7-DoF robot arm.
For the real robot, we use a motion capture system (MoCap) to estimate the pose of objects and assess the quality of the learned controller.

The robot setup is shown in Fig. \ref{fig:method} ("environment real/sim").  
We use a 7-DoF Franka Emika Panda robot arm with a custom spherical end-effector attached to interact with the object.
The task at hand is to push the cuboid on the planar surface of a table to move it into a target pose configuration.
The weighting parameters of the MPC cost function are empirically tuned to $\lambda_1=2\cdot10^{-2}, \lambda_2=5\cdot10^{-4}, \lambda_3=10^{-5}$.
To synchronize the state of the internal dynamics model to the real world ahead of every MPC execution, the end-effector and object poses $\mathbf{q}_r, \mathbf{q}_o$ are obtained as part of the world state $\mathbf{s}_w$ from MoCap at 240Hz.
The corresponding end-effector velocities~$\mathbf{v}_r$ are provided at 1kHz by the low-level robot interface and transformed from the end-effector into the world frame.
With a planning horizon of 0.6\,s, the MPC achieves on average a 10\,Hz replanning frequency for the real robot using a PC with i9-10900X CPU. 
For the MPC optimization we set $N=50$ and $C=2$. 
For the parameter optimization we set $Q=32$ and $C=2$.
The simulation setup is shown in Fig. \ref{fig:method} ("environment real/sim"). 
We model the robot end-effector as a free-floating sphere primitive. 
The object to be interacted with is modeled as a cuboid shape primitive.
As in simulation, environment execution can be paused during planning of the next trajectory keypoints, arbitrary planning frequencies can artificially be imitated. 
We set the frequency to 20\,Hz. 
For the MPC optimization we set $N=128$ and $C=4$. For the parameter optimization we set $Q=32$ and $C=2$.

\begin{figure}[tbp] 
    \centering
    \vspace*{1.2ex}
\includegraphics[height=0.22\linewidth]{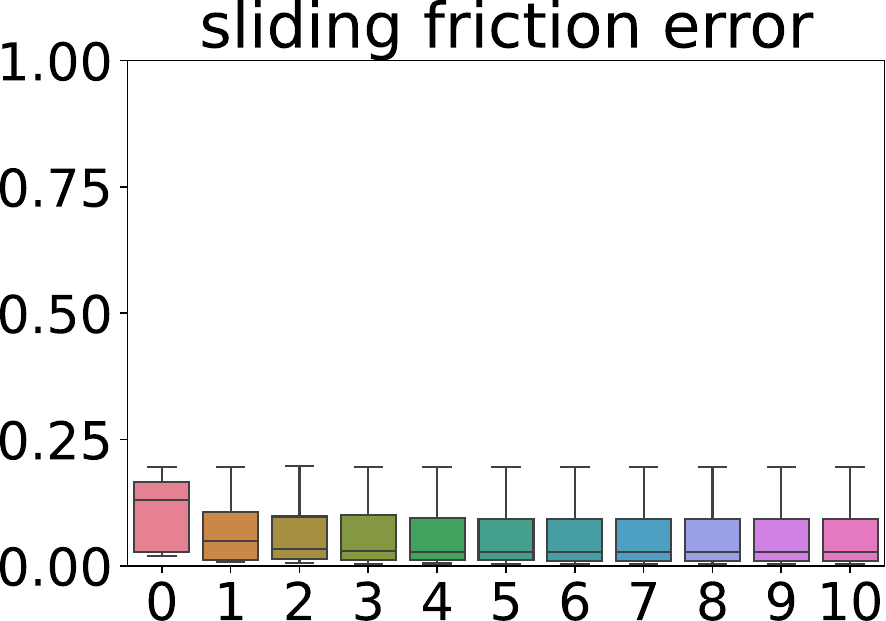} \hfill
    \includegraphics[height=0.22\linewidth]{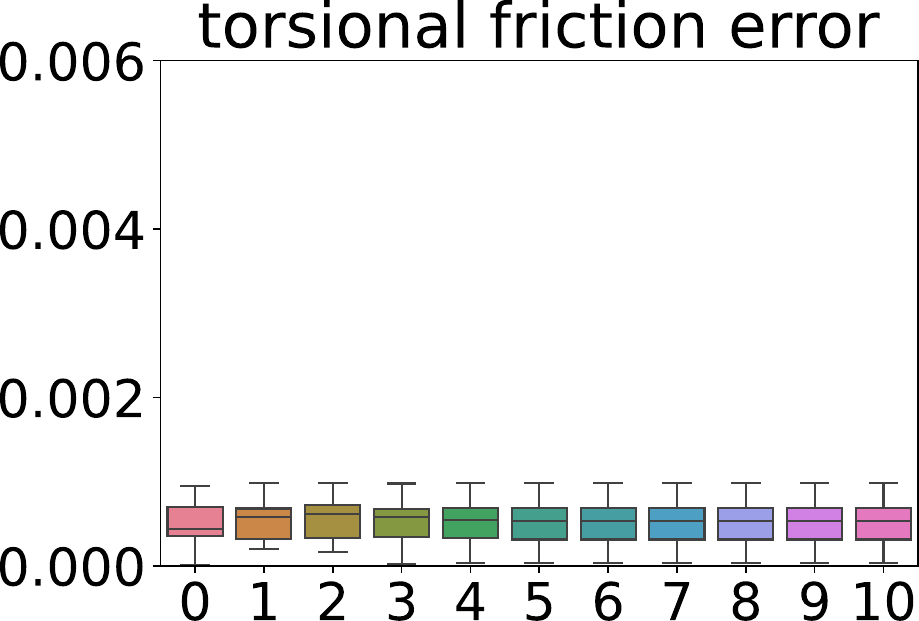}\hfill
    \includegraphics[height=0.22\linewidth]{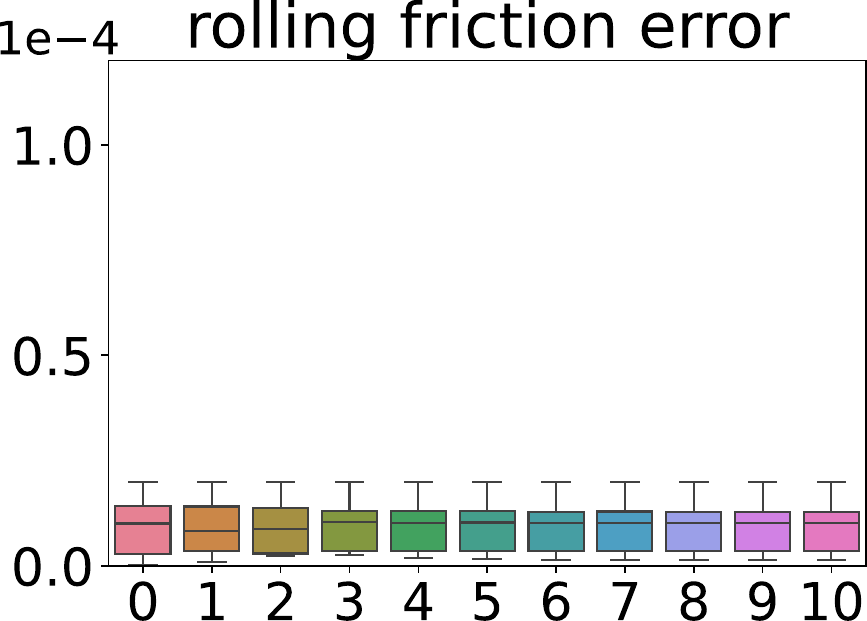}\\
    \vspace*{1ex}
    \includegraphics[height=0.25\linewidth]{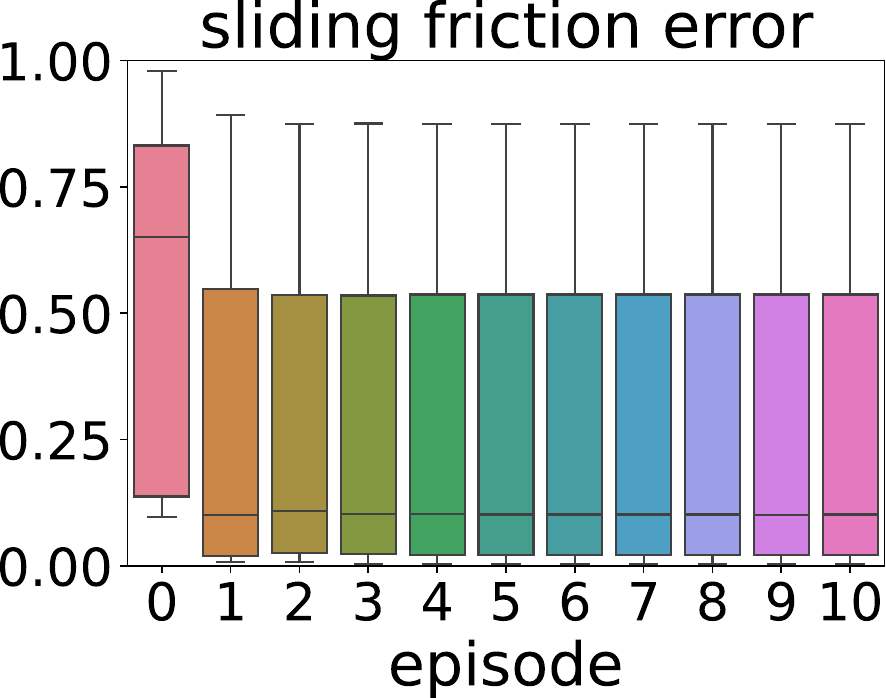}\hfill
    \includegraphics[height=0.25\linewidth]{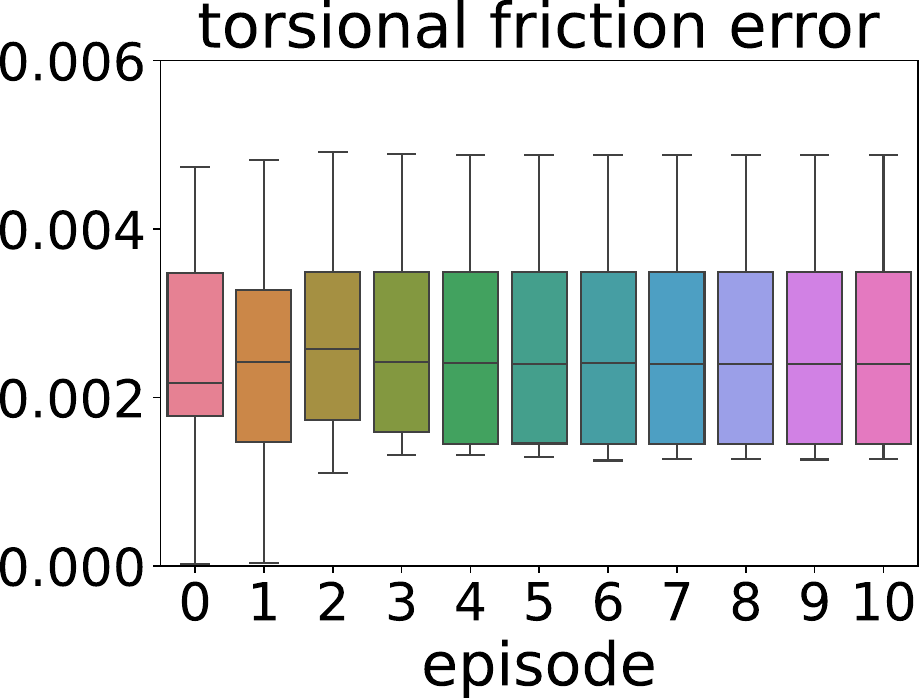}\hfill
    \includegraphics[height=0.25\linewidth]{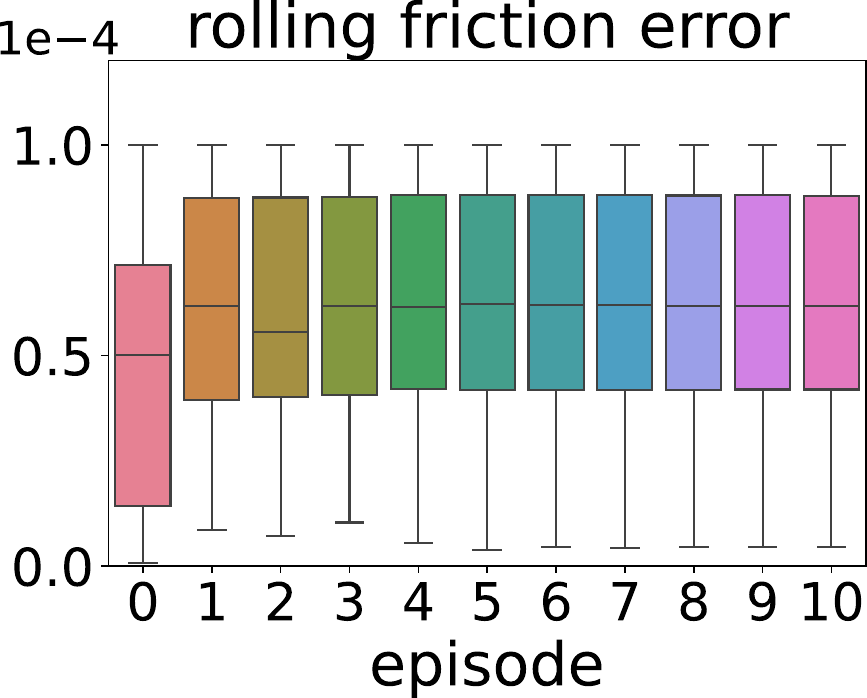}
    \caption{Parameter estimation errors of sliding, torsional and rolling friction for $\Delta=0.2$ (top) and $\Delta=1$ (bottom) during optimization over 10 episodes. Boxes are bounded by upper and lower quartiles. Initial error at "episode"~$0$.}
\label{fig:param_errors}
\end{figure}

\begin{figure}[tbp]
    \centering
    \includegraphics[height=0.26\linewidth]{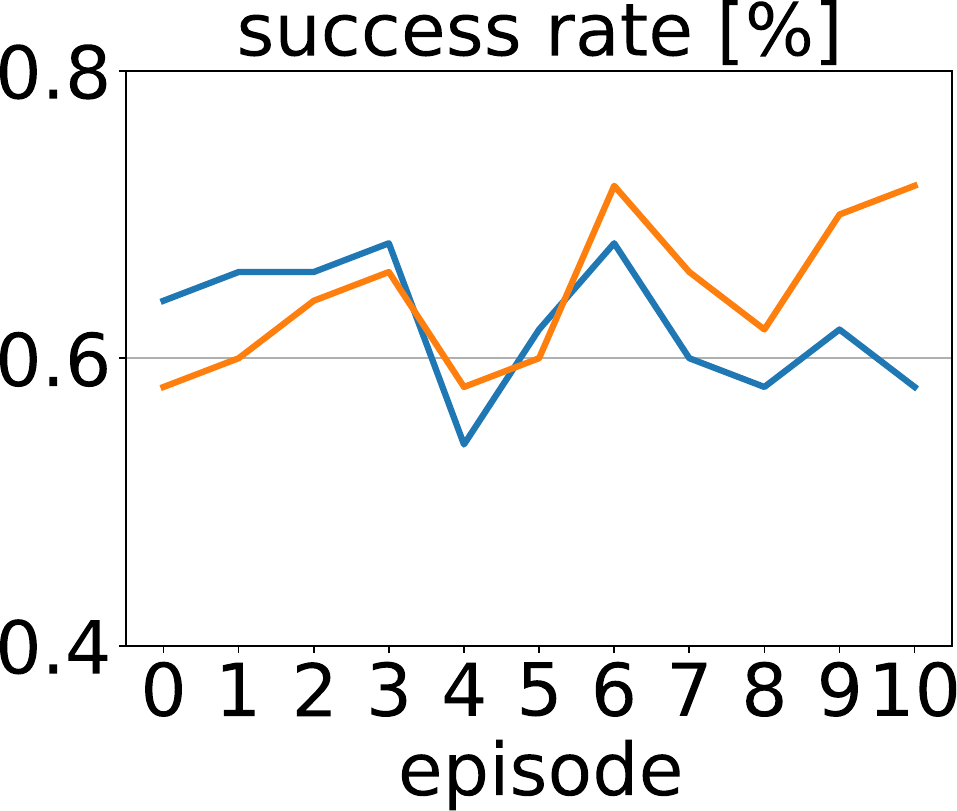} \hfill
    \includegraphics[height=0.26\linewidth]{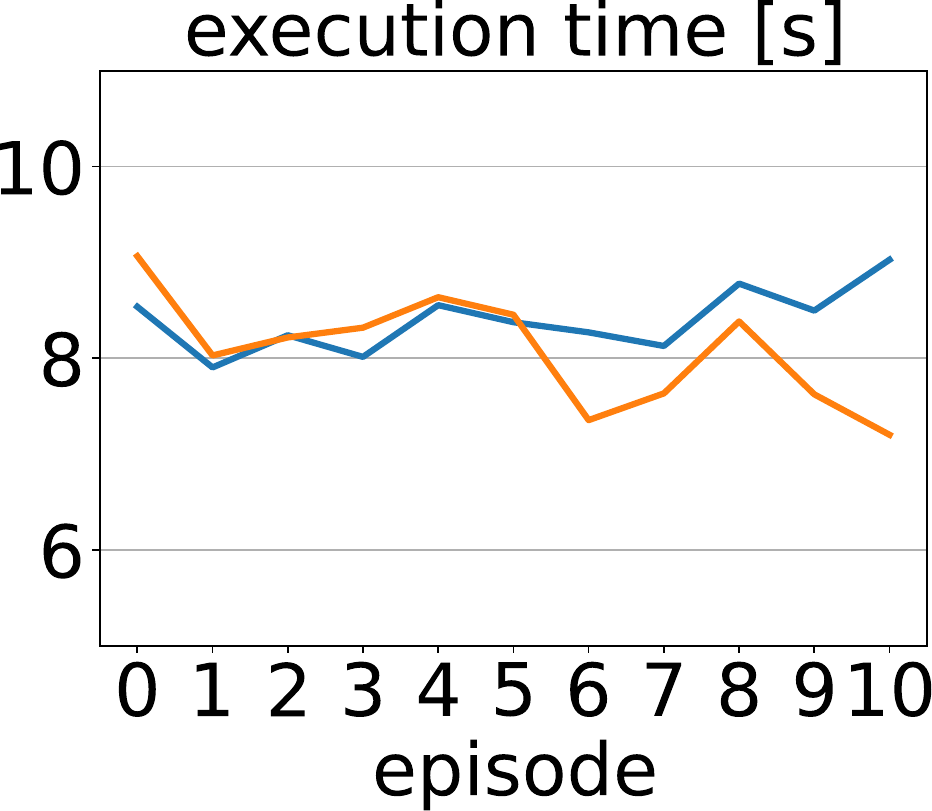} \hfill
    \includegraphics[height=0.26\linewidth]{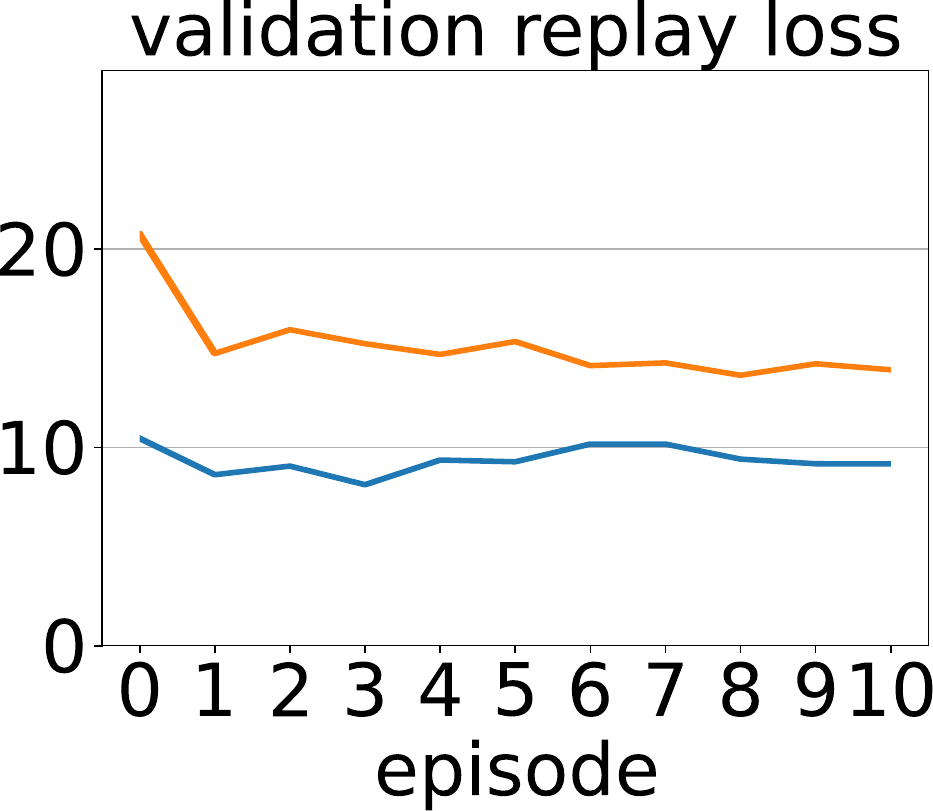}\hfill
    \caption{Success rate, mean task execution time, and validation replay loss for $\Delta = 0.2$ (blue) and $\Delta = 1$ (orange).}
\label{fig:sim_eval}
\end{figure}

\begin{figure}
   \centering
   \vspace*{1.2ex}

\includegraphics[width=0.99\linewidth]{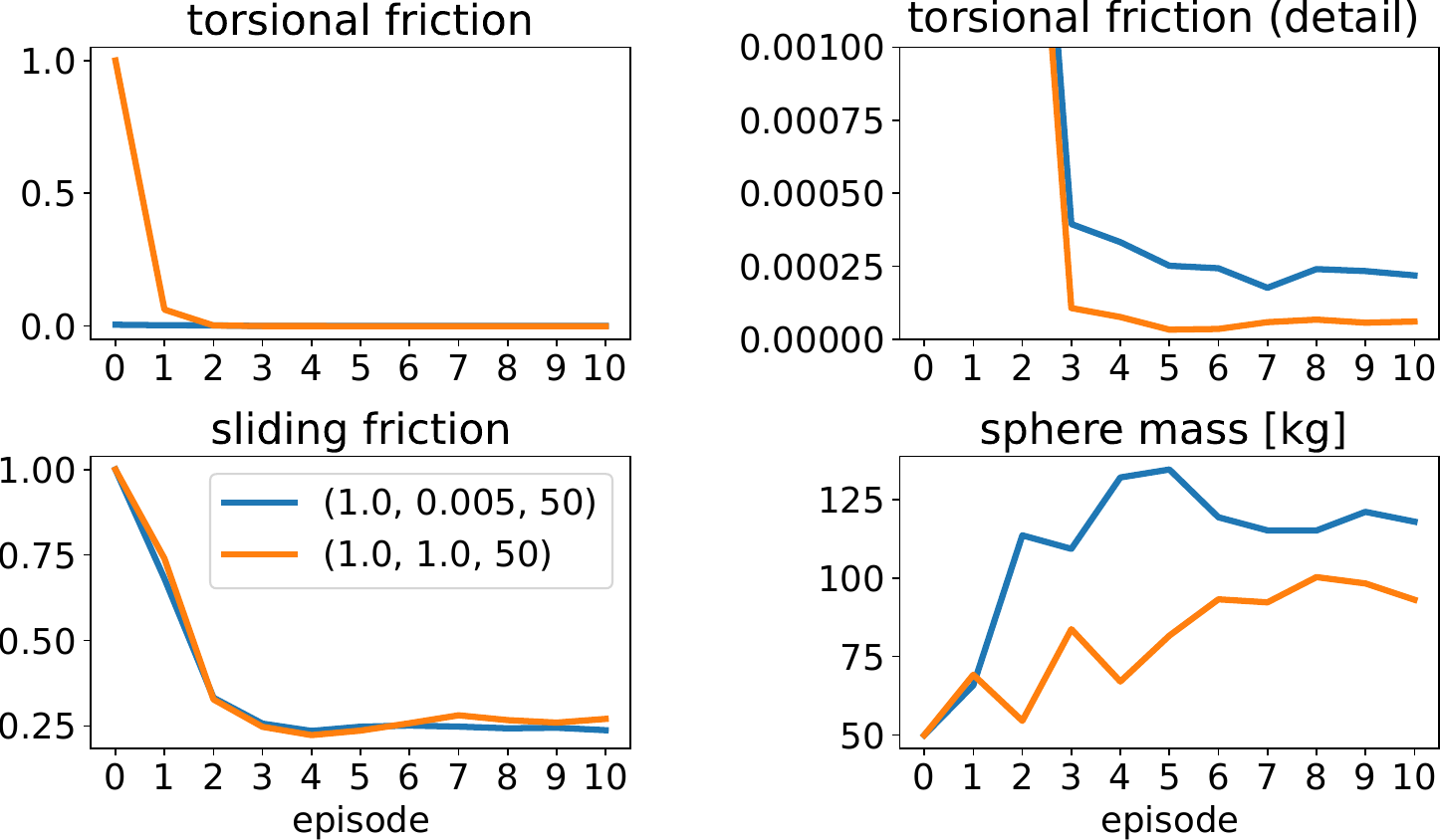}
   \caption{Model parameters over optimization episodes, starting from two initial sets (orange: $( \theta_{s}, \theta_{t}, \theta_m) = (1.0, 1.0, 50)$, blue: $( \theta_{s}, \theta_{t}, \theta_m) = (1.0, 0.005, 50)$).}
\label{fig:param_iters}
\end{figure}

\begin{figure}[tb]
   \centering
   \includegraphics[width=\linewidth]{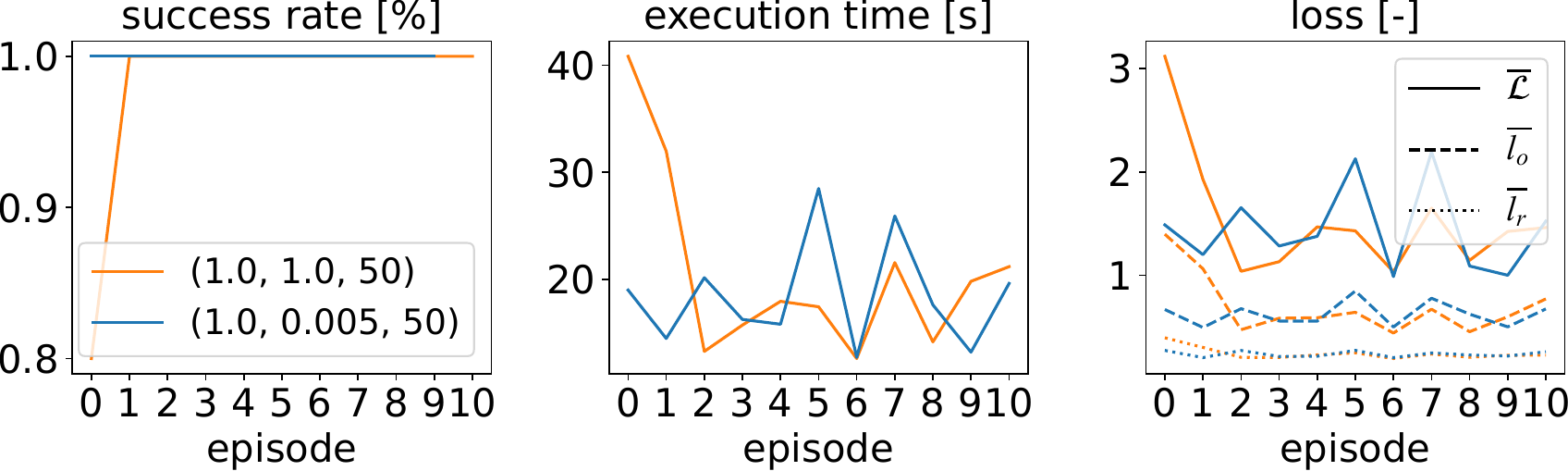}
   \caption{Success rate~$S$, avg. execution time~$\overline{t}$, avg. MPC loss over the executed trajectories~$\overline{\mathcal{L}}$, avg. object trajectory length~$\overline{l}_{o}$ and avg. robot end-effector trajectory length~$\overline{l}_{r}$ for initial parameter sets~$( \theta_{s}, \theta_{t}, \theta_m) = (1.0, 1.0, 50)$ (orange) and~$( \theta_{s}, \theta_{t}, \theta_m) = (1.0, 0.005, 50)$ (blue).}
\label{fig:metrics}
\end{figure}

\begin{figure*}[tbp!]
    \centering
\vspace*{1.2ex}\includegraphics[width=0.99\textwidth]{src/new_all_rectangles_init14_epoch0_run1_new.pdf}\\
    \vspace{1ex}
    \includegraphics[width=0.99\textwidth]{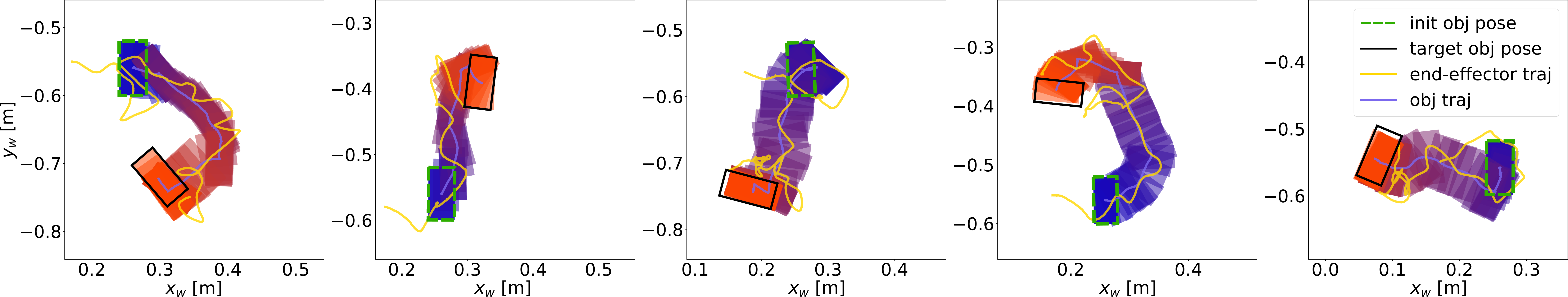}\\
    \vspace{1ex}
    \includegraphics[width=0.15\textwidth]{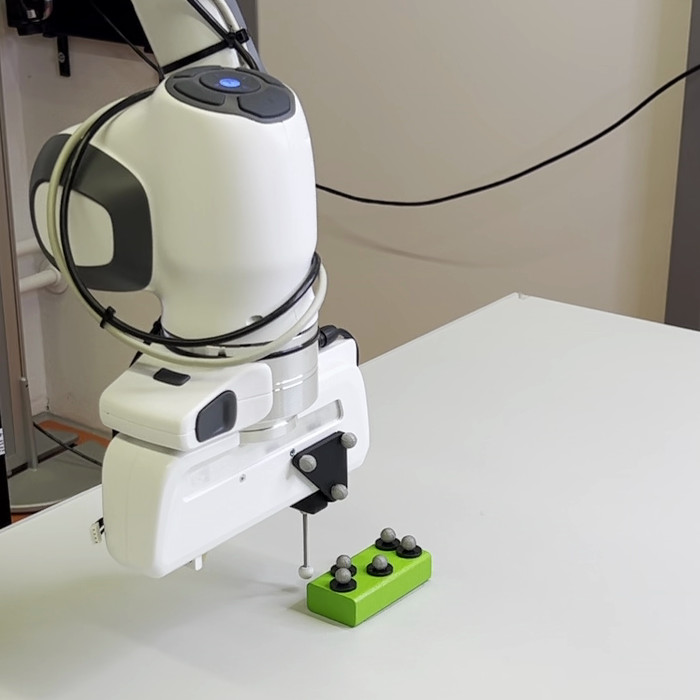}
    \includegraphics[width=0.15\textwidth]{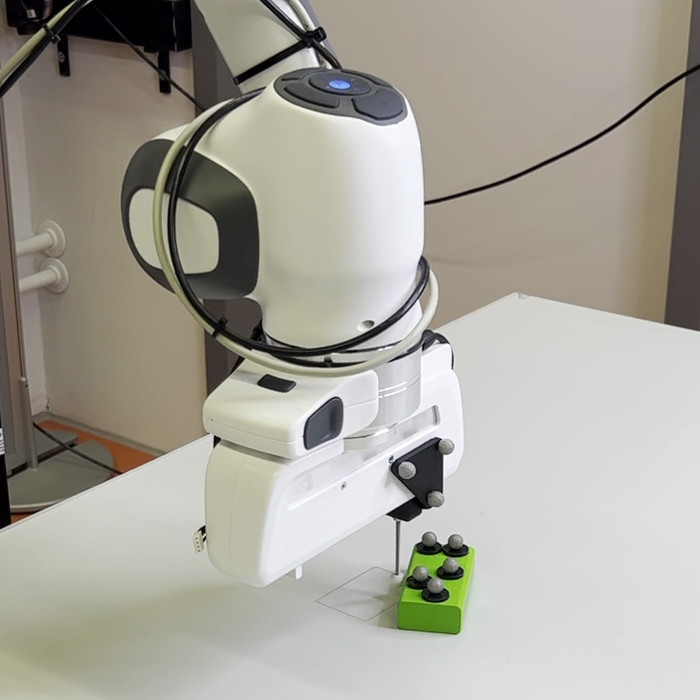}
    \includegraphics[width=0.15\textwidth]{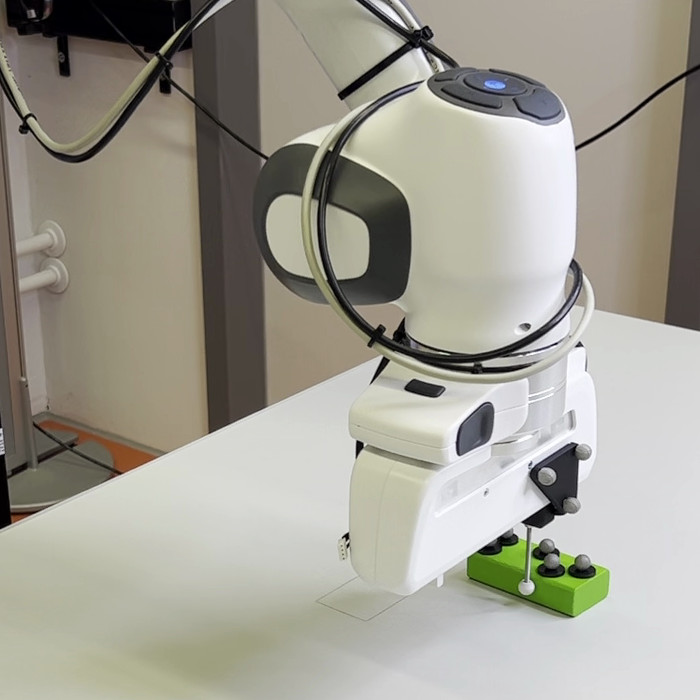}
    \includegraphics[width=0.15\textwidth]{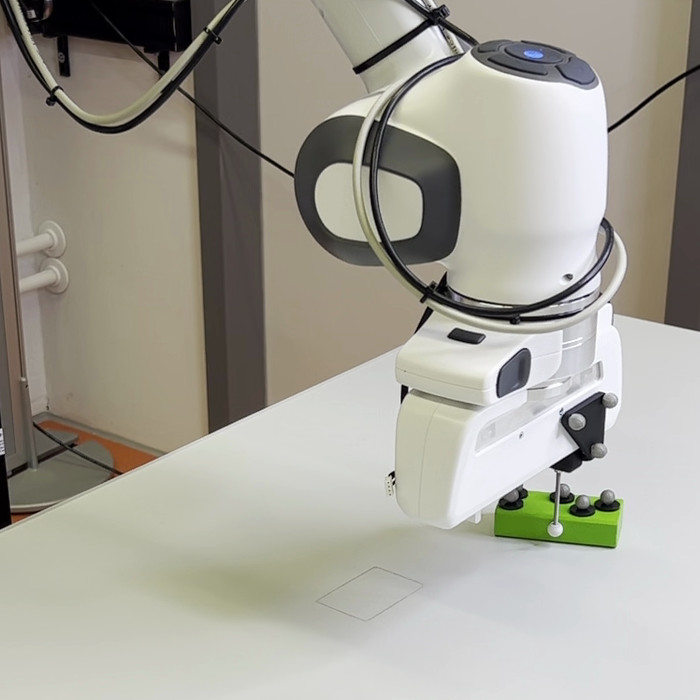}
    \includegraphics[width=0.15\textwidth]{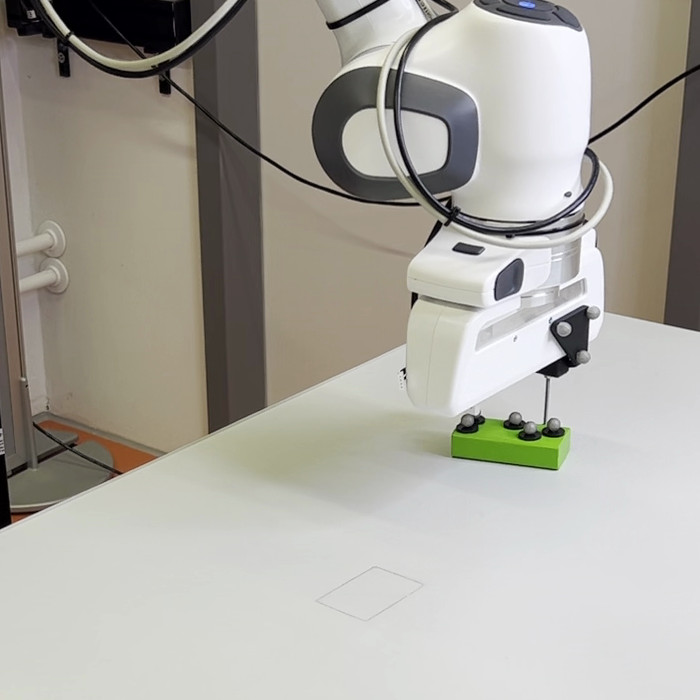}
    \includegraphics[width=0.15\textwidth]{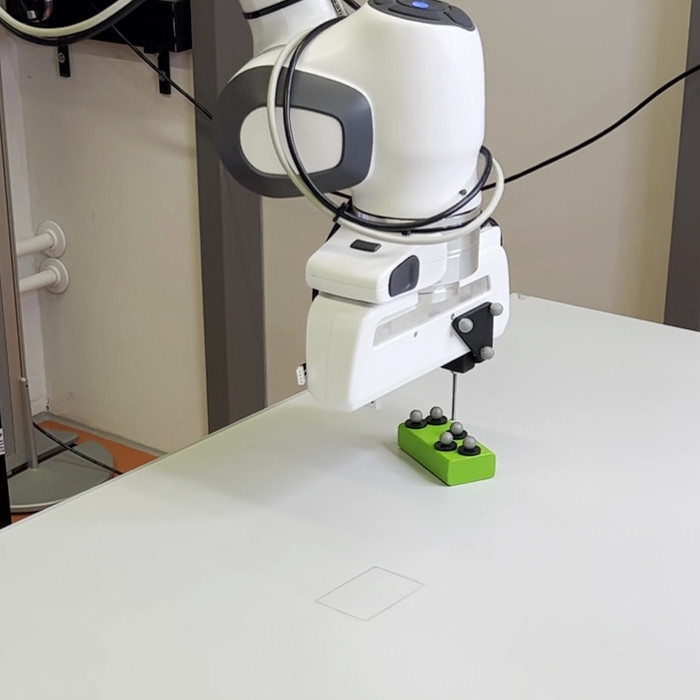}
    \caption{Object and end-effector trajectories from five object pushing tasks on the real robot. Upper row: MPC performance using the initial model parameters~$( \theta_{s}, \theta_{t}, \theta_m) = (1.0, 1.0, 50)$ for the dynamics model. Middle row: MPC performance using the model parameters obtained after 6 optimization episodes $( \theta_{s}, \theta_{t}, \theta_m)=(2.58\cdot10^{-1}, 3.56\cdot10^{-5}, 9.32\cdot10^{1})$.
    Bottom row: Object pushing task 4 executed on the robot with the parameter set obtained after 6 optimization episodes.
    }
    \label{fig:all_rec}
\end{figure*}

We evaluate the performance of the MPC based on a set of executed task trajectories $\mathbf{S}=(\mathbf{S}_{1}, \dots, \mathbf{S}_{H})$. 
The average success rate is
    $S=\frac{1}{H} \sum_{j=1}^H \boldsymbol{1}_{\text{success},j}$,
where~$\boldsymbol{1}_{\text{success},j}$ indicates success according to the first terminal condition in Sec.~\ref{sec:mpcnonprehense} for $\epsilon = 10^{-3}$ in simulation and for $\epsilon = 10^{-4}$ in real-world experiments. 
The mean execution time is
    $\overline{t}=\frac{1}{H} \sum_{j=1}^H t_{\mathit{terminal},j}$,
where~$t_{\mathit{terminal},j}$ is the termination time for task~$j$.
The average object trajectory loss
    $\overline{\mathcal{L}}_o(\mathbf{S})
    =\frac{1}{H}\cdot 10^{-4} \sum_{j=1}^H \sum_{t=0}^T
    \left( \mathcal{L}_{pos}(\mathbf{s}_{j,t})+\lambda_1\mathcal{L}_{rot}(\mathbf{s}_{j,t}) \right)$
is computed from the positional~$\mathcal{L}_{pos}$ and rotational error~$\mathcal{L}_{rot}$ as in Sec.~\ref{sec:mpcnonprehense}.
We also compute the average trajectory lengths by
    $\overline{l}_{o/r}(\mathbf{S})
    =\frac{1}{H} \sum_{j=1}^H \sum_{t=0}^{T-1}
    \left \| \overline{\mathbf{y}}_{o/r,j,t+1} - \overline{\mathbf{y}}_{o/r,j,t} \right \|_2$,
where in simulation~$\overline{\mathbf{y}}_{o/r}$ are the trajectories of object and robot end-effector, respectively. In the real-world experiments, they denote MoCap-recorded trajectories smoothed with a Gaussian kernel over time ($\sigma=100$ at 1\,kHz).

\subsection{Simulation Results}
In simulation, we can assess performance of recovering ground truth physical parameters by our method.  
We can also directly replay the forces applied to the sphere.
We perform two experiments for different levels of uncertainty in the initial parameter range. 
In our experiments we optimize a subset of physical parameters consisting of sliding friction $\theta_s$, torsional friction $\theta_t$ and rolling friction $\theta_r$, keeping the sphere mass $\theta_m$ at its ground truth value.
In all experiments, we set the ground truth friction parameters to the default \textit{MuJoCo} settings ($(\widehat{\theta}_s, \widehat{\theta}_t, \widehat{\theta}_r) = (1, 0.005, 0.0001)$).
CEM initially samples parameters from a normal distribution which depends on prior knowledge about the parameter range.
In each run, the initial mean~$\mu_{\theta, 0} \in \mathcal{U}(a, b)$ of each parameter's distribution is sampled from a uniform distribution around the ground truth value~$\widehat{\theta}$ with lower and upper bounds~$a = \widehat{\theta} - \Delta \cdot \theta$ and $b = \widehat{\theta} + \Delta \cdot \theta$, respectively.
The size of the sampling interval $\Delta \cdot \theta$ is chosen according to assumed prior knowledge about the parameter range.
We use settings of $\Delta \in \left\{ 0.2, 1.0 \right\}$ in the experiments.
The standard deviation of the distribution is set to $(\widehat{\theta}+\Delta \cdot \widehat{\theta}-\max(10^{-7}, \widehat{\theta}-\Delta \cdot \widehat{\theta}))^2/16$.

\subsubsection{Parameter Optimization}
For each experiment, we perform ten runs with independently sampled initial parameter means~$\mu_{\theta, 0}$.
To obtain samples of robot-object-interactions, an object pushing task is executed in each episode with the current set of best parameters. For task completion, we set a time limit of 5\,s.
Fig. \ref{fig:param_errors} shows the parameter estimation errors over ten optimization episodes. For both $\Delta$, the sliding friction coefficient error drops after the first episode and converges afterwards. Torsional and rolling friction errors stay similar for $\Delta = 0.2$ and $\Delta = 1$ over the ten episodes.

\subsubsection{MPC Performance}
Each run is independently evaluated on five tasks with different object target positions and orientations in terms of success rate and execution time.
Fig.~\ref{fig:sim_eval} shows mean values over ten independent experiment runs after each episode.
Due to the large variability how MPC executes tasks, these results are difficult to interpret.
For $\Delta = 1$, a small improvement seems to be visible over the episodes in both metrics.
Fig.~\ref{fig:sim_eval} also shows avg. loss $\mathcal{L}(\boldsymbol{\theta})$ for replays with the optimized parameters after each episode on five validation rollouts obtained by task execution with ground truth parameters.
Here, $\Delta = 1$ improves performance in the first episode and seems to converge afterwards.

\subsection{Real-World Results}

We evaluate our incremental few-shot adaptation approach on the real robot system by sequentially performing object pushing tasks to random goal locations, adding the trajectory to the replay buffer, and optimizing the model parameters for the next iteration. As no object rolling motion can be observed, we optimize the sliding friction $\theta_s$ and the torsional friction $\theta_t$. In comparison to the simulation the real-world end-effector differs from the dynamics model end-effector. Therefore we additionally optimize the end-effector sphere mass $\theta_m$.
Our approach does not require high accuracy end-effector contact force measurements,
instead we replay end-effector velocities in the simulation.
The parameter set of each iteration is evaluated with a different set of 5 object pushing tasks which are the same for all iterations.
Each task requires the spherical robot end-effector to push the cuboid object from a fixed initial pose~$(\mathbf{y}^0_{o}, \phi^0_{o})$ to a different target pose~$(\mathbf{y}_{o}^*, \phi_{o}^*) \in \mathbb{R}^3$.
All tasks start from the same initial pose so that tasks do not dependent on the achieved object pose of the previous task, thus allowing to compare the evaluation performance of episodes across runs.

We assess two parameter sets.
The first is an offset initialization, with both the sliding and torsional friction initialized to one, while the sphere mass is empirically set to 50\,kg.
The second parameter set is the default configuration of MuJoCo, where the sliding friction is set to one and the torsional friction is set to 0.005, while the sphere mass is set to 50\,kg.
Fig.~\ref{fig:param_iters} shows the evolution of the parameters in both runs.
Index~$i$ denotes that the parameter set obtained after the~$i$-th optimization iteration is evaluated.
Note that ground truth values are unknown.
Fig.~\ref{fig:metrics} shows quantitative results for both parameter initializations.
With the initial offset parameters, MPC performance is worse in all metrics compared to later optimization episodes.
With the parameters after the 6th optimization episode ($i=6$), MPC needs on average only approx. one-third time to complete a task. 
Also, the end-effector and object trajectories are shorter.
A visual comparison of the object pushes with the initial (top row) and 6th (bottom row) parameter set is displayed in Fig.~\ref{fig:all_rec}.
Strong improvements towards previous iterations can be observed in the first two episodes.
After that, while success rate stays high, the performance metrics fluctuate and no clear further trend of improvement is visible presumably due to the inherent stochasticity of task execution. 
Torsional friction and sphere mass still adapt to a specific level and show smaller changes after the 6th episode.
It can be seen that the motions have lowest avg. execution time in all tasks for the 6th episode parameters. 
With the initialization parameters set to default MuJoCo values, no clear trend in the performance of the MPC task execution over the iterations can be observed, presumably since the initial torsional friction is already close to its optimized value and due the stochastic variability of task execution. 
The parameters converge to similar values for both initializations (see Fig.~\ref{fig:param_iters}).
For both initial parameter sets the controller stopped during multiple episodes due to exceeding joint acceleration or velocity limits of the robot (episodes 0, 1 and 4 of the first initial parameter set and episode 0 of the second initial parameter set). 
These episodes were repeated.
This is a limitation of our Cartesian space minimum snap trajectory controller which could be addressed in future work,
f.e., by checking joint limits along trajectories during planning or planning in joint space.

\section{CONCLUSION}
In this paper we propose a novel MPC-based approach that makes use of a rigid-body physics simulation as a dynamics model and adapts the model incrementally. 
The MPC combines sample-efficient optimization with a minimum snap trajectory formulation for real-time capability.   
Our method applies incremental sampling-based optimization of the model parameters to align the simulation with actual interaction rollouts collected during task execution.
We analyze our MPC approach including parameter optimization in simulation and on a real robot setup.
In simulation, our approach improves model parameters, especially for larger offset initializations, to better align with replay rollouts. 
For the real robot, our incremental parameter adaptation approach leads to improved MPC performance in terms of success rate, execution time, and trajectory length from a few real-world episodes for an offset parameter initialization.
In future work, we plan to develop our approach further for a larger diversity of tasks. To this end, the complexity of the dynamics model could be increased and the model could be adapted to changing environment conditions.

\balance

\bibliographystyle{IEEEtranS}
\bibliography{refs.bib}

\end{document}